\documentclass[11pt,a4paper]{article}
\usepackage[final]{acl}
\usepackage{times}
\usepackage{latexsym}
\usepackage[T1]{fontenc}
\usepackage[utf8]{inputenc}
\usepackage{microtype}
\usepackage{inconsolata}
\usepackage{graphicx}
\usepackage{booktabs}
\usepackage{amsmath}

\title{Can LLMs Catch a Rigged Backtest?\\
A Clean-Control Calibration Benchmark}

\author{Makar Ulesov\textsuperscript{1} \quad
Vladislav Smirnov\textsuperscript{1} \quad
Omar Ibrahim\textsuperscript{1} \quad
Arsenii Bobovnikov\textsuperscript{2} \\
\textsuperscript{1}Mohamed bin Zayed University of Artificial Intelligence (MBZUAI) \\
\textsuperscript{2}School of Computing and Information, University of Pittsburgh}

\hypersetup{
  pdftitle={Can LLMs Catch a Rigged Backtest? A Clean-Control Calibration Benchmark},
  pdfauthor={Makar Ulesov; Vladislav Smirnov; Omar Ibrahim; Arsenii Bobovnikov}
}

\begin{document}
\maketitle
\begin{abstract}
Backtest auditing is a calibration problem: high flaw recall is not useful when the model falsely flags matched clean strategies. We build a 96-item paired benchmark in which every flawed backtest has a clean control that holds strategy, dates, code style, labels, and reporting scaffold fixed while changing one methodology detail. A deterministic scorer separates flaw recall, clean-control false positives, evidence localization, and fix relevance. Over 1440 cached audits from four text endpoints, the primary DeepSeek auditor reaches 100.0\% closed and clean-aware code recall, but open prompts over-flag 93.8\% of clean code controls, and clean-aware all-three specificity is 87.5\% even where recall saturates. A clean-aware warning drops DeepSeek code false positives from 20.8\% (95\% CI 11.7--34.3) to 0.0\% (0.0--7.4) at unchanged recall, while the budget anchor still flags 38/48 clean controls under the same prompt. Reporting recall alone would rank three of these four models identically; reporting the clean-control rate separates them by 79 points.
\end{abstract}

\section{Calibration as Much as Flaw-Finding}
Backtest review is a calibration problem as much as a flaw-finding problem. A useful LLM auditor must detect look-ahead, survivorship, leakage, p-hacking, and execution errors \emph{while also} letting a clean strategy pass; closed-set recall alone is misleading because false positives add review threads and can obscure the one backtest that actually needs repair.

\begin{figure*}[t]
  \centering
  \includegraphics[width=\textwidth]{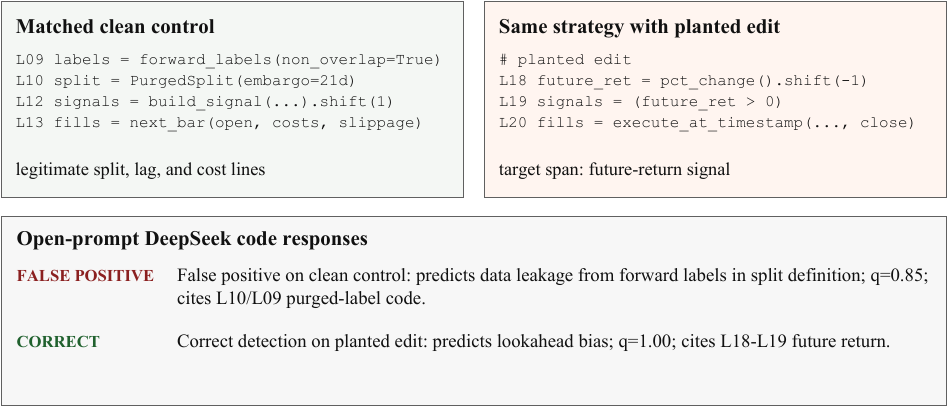}
  \caption{Paired clean-control design. Left: legitimate split, lag, and cost lines in a clean control. Right: the same strategy with a planted look-ahead edit. The same open code audit correctly flags the flawed edit but falsely accuses the matched clean control; without a matched control, the false accusation would look like a plausible audit insight.}
  \label{fig:pair}
\end{figure*}

We therefore make calibration the object of study with a controlled \emph{paired} benchmark. Each flawed item has a matched clean control that keeps strategy, dates, code style, labels, and reporting scaffold fixed while changing one methodology detail, which separates flaw recognition from a high suspicion prior (Figure~\ref{fig:pair}). The benchmark does not certify trading systems; it isolates one review skill.

Prior quantitative-finance work owns backtest overfitting, data snooping, and survivorship bias \citep{white2000,harvey2016,bailey2014,lopez2018}. Prior LLM-finance and LLM-review benchmarks own backtest generation, look-ahead mitigation, issue-level critique, and LLM-assisted review \citep{wang2026,li2026,chaudhuri2026,ho2026,zurawicki2026}. This work differs by jointly evaluating matched clean controls, prose and code surfaces, canonical flaw labels, and deterministic evidence and fix scoring.

\section{Benchmark and Validator}
\paragraph{Items.}
The 96-item set has 48 clean/flawed pairs (48 flawed, 48 clean controls), 6 flawed examples per flaw class, three difficulty strata, and two surfaces per item (prose and code) over $|\mathcal{K}|=8$ canonical flaws: look-ahead, survivorship, data snooping, full-sample and overlapping-label leakage, unrealistic fills/costs, regime overfit, and p-hacked stop/rule variants, plus a clean label. Clean controls are not merely the absence of an edit. They contain the protective constructs a careful quantitative researcher would write---purged and embargoed splits, explicit one-way costs, half-spread slippage, non-overlapping labels---so a model that pattern-matches on the presence of those constructs will misfire in both directions.

\paragraph{Conditions.}
The prompt grid crosses \{open, closed, clean-aware\} with \{prose, code\}. Open omits the taxonomy, closed includes it, and clean-aware adds that clean controls are common. The primary auditor is DeepSeek V4 Flash at temperature 0; the code-slice ladder adds \texttt{gpt-4o-mini} (budget anchor), \texttt{gemini-2.5-\allowbreak flash-lite}, and \texttt{gpt-4.1-mini}. All 1440 item-condition calls parse after a token-cap rerun, at a parse rate of 100.0\%.

\paragraph{Scoring.}
The deterministic validator scores five quantities separately: clean-control false-positive rate $F_c$, with review burden $C_c=100F_c$ unnecessary threads per 100 clean submissions; evidence credit, where the cited span must hit the planted target lines; fix credit, where a class-specific keyword earns the point and generic advice earns nothing; all-three specificity, meaning simultaneously class-, evidence-, and fix-correct; and a utility $U_c(\lambda)=\bar{R}_c-\lambda F_c$ with $\lambda\in\{0.25,1,2\}$. Binomial intervals are Wilson score intervals at 95\%. The dataset seed is \texttt{20260612}.

\section{Results: Recall Is Easy, Calibration Is Not}
\label{sec:results}

\begin{table}[t]
\centering
\caption{Primary DeepSeek prompt-by-surface confusion matrix. Flawed cells are true positives/false negatives; clean cells are false positives/true negatives. Threads are expected unnecessary reviews per 100 clean submissions.}
\label{tab:primary}
\footnotesize
\setlength{\tabcolsep}{1.5pt}
\begin{tabular*}{\columnwidth}{@{}l@{\extracolsep{\fill}}rrrrr@{}}
\toprule
Condition & TP/FN & FP/TN & $R$ & FPR & Threads \\
\midrule
open prose       & 39/9  & 18/30 & 81.2  & 37.5 & 37.5 \\
open code        & 35/13 & 45/3  & 72.9  & 93.8 & 93.8 \\
closed prose     & 48/0  & 1/47  & 100.0 & 2.1  & 2.1  \\
closed code      & 48/0  & 10/38 & 100.0 & 20.8 & 20.8 \\
clean-warn prose & 48/0  & 0/48  & 100.0 & 0.0  & 0.0  \\
clean-warn code  & 48/0  & 0/48  & 100.0 & 0.0  & 0.0  \\
\bottomrule
\end{tabular*}
\end{table}

Table~\ref{tab:primary} is the primary prompt-by-surface confusion matrix. Open code audits detect 35/48 flawed items but create 45 unnecessary review threads per 48 clean controls, i.e.\ 93.8 per 100 clean submissions. Closed and clean-aware prompts both reach 100.0\% code recall, but only the clean-aware prompt also removes DeepSeek clean false positives on this paired set, dropping code FPR from 20.8\% to 0.0\% and prose FPR from 2.1\% to 0.0\% while preserving 100.0\% recall.

The clean-control drop is larger than sampling noise. On code the closed prompt flags 10/48 clean controls, a rate of 20.8\% with a 95\% Wilson interval of 11.7--34.3\%; the clean-aware prompt flags 0/48, an interval of 0.0--7.4\%. Those intervals do not overlap. The corresponding recall does not move: both prompts sit at 48/48.

\paragraph{Perfect recall is not complete audit quality.}
Clean-aware code all-three specificity is only 87.5\% (95\% CI 75.3--94.1), decomposing into 100.0\% evidence localization and 87.5\% fix relevance. The model points at the right lines in every flagged item and still proposes an off-class remedy in six of them. Prose is the easier surface here, at 100.0\% all-three specificity under the same prompt. A single audit score would hide the fact that the residual failure sits entirely in the fix column.

\begin{table}[t]
\centering
\caption{Clean-aware code slice, four endpoints, 48 flawed and 48 clean items each. FPR intervals are 95\% Wilson. A3 is all-three specificity; $U_\lambda$ is $\bar{R}-\lambda F$.}
\label{tab:models}
\footnotesize
\setlength{\tabcolsep}{1pt}
\renewcommand{\arraystretch}{1.08}
\begin{tabular*}{\columnwidth}{@{}l@{\extracolsep{\fill}}rrrrr@{}}
\toprule
Model & $R$ & FPR (95\% CI) & A3 & $U_1$ & $U_2$ \\
\midrule
\shortstack[l]{DeepSeek V4\\Flash} & 100.0 & 0.0 (0.0--7.4) & 87.5 & 1.00 & 1.00 \\
\shortstack[l]{Gemini 2.5\\Flash Lite} & 87.5 & 0.0 (0.0--7.4) & 85.4 & 0.88 & 0.88 \\
GPT-4.1 mini & 100.0 & 16.7 (8.7--29.6) & 97.9 & 0.83 & 0.67 \\
GPT-4o mini & 100.0 & 79.2 (65.7--88.3) & 75.0 & 0.21 & $-0.58$ \\
\bottomrule
\end{tabular*}
\end{table}

\paragraph{Calibration varies by model.}
Table~\ref{tab:models} separates the four endpoints on a column that recall cannot see. DeepSeek and Gemini have 0/48 clean false positives, GPT-4.1 mini has 8/48, and the budget anchor GPT-4o mini has 38/48, despite all four keeping high recall. Under $U_c(\lambda)$, DeepSeek clean-aware code has $U_{0.25}=U_1=U_2=1.00$, while GPT-4o mini clean-aware code falls from $U_{0.25}=0.80$ to $U_1=0.21$ to $U_2=-0.58$ despite perfect recall. Three of the four models tie at 100.0\% recall and span a 79-point range on the clean-control rate.

The rule baseline is a sanity anchor only, at 25.0\% code recall and 100.0\% code clean FPR, since clean controls contain protective phrases such as purges and realistic slippage. False positives are not parser noise. Most name an incorrect flaw class and point at a legitimate safeguard: one flags \texttt{unrealistic\_fills\_costs} while citing a line that sets a 10~bp one-way cost with half-spread slippage, and another flags \texttt{overlapping\_label\_leakage} while citing non-overlapping labels combined with a purged walk-forward split at a 21-day embargo. Self-reported confidence does not rescue the ranking either. Among the 281 audits in the top confidence bin (above 0.85, mean 0.939), 35.0\% of the clean-control judgments are still false positives.

\begin{figure*}[t]
\centering
\includegraphics[width=\textwidth]{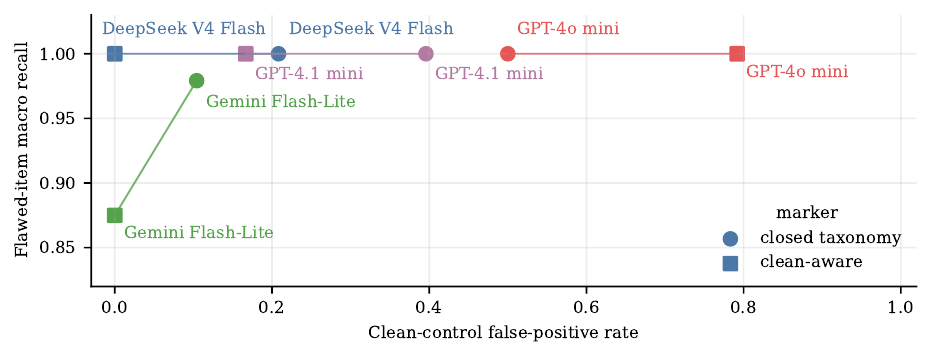}
\caption{Clean-control false-positive rate by model and prompt mode on the code surface. The vertical spread at the clean-aware prompt is the quantity that recall-only reporting discards: every model in that column is at or near ceiling recall.}
\label{fig:models}
\end{figure*}

\paragraph{Where open prompts fail.}
The open-prompt deficit is concentrated rather than spread. On code, the primary auditor recovers 6/6 look-ahead, survivorship, and data-snooping items but only 1/6 full-sample preprocessing and 2/6 regime overfit; on prose the same two classes fall to 0/6 and 4/6. Difficulty strata do not explain it, since open code recall is 12/16 obvious, 11/16 moderate, and 12/16 subtle. Without the taxonomy the model finds the flaws that have familiar names and misses the ones that live inside a preprocessing step. Appendix~\ref{app:classes} gives the full per-class breakdown.

\paragraph{What the rates cost a reviewer.}
The false-positive column converts directly into review load, which is why we report $C_c=100F_c$ alongside $F_c$. A team that runs the open code prompt over a submission queue of 100 clean strategies opens 93.8 unnecessary threads. The same team running the clean-aware prompt with the primary auditor opens none, and running it with the budget anchor opens 79.2. Recall is identical for the second and third of these, so the entire difference in operating cost is invisible to a recall-only report.

The asymmetry also runs the other way. Under the open prompt the primary auditor misses 13 of 48 flawed code items while raising 45 unsupported flags, so the flagged pile contains more clean strategies than rigged ones. An auditor in that regime is worse than uninformative for triage, because the reviewer who works the queue in order spends most of the budget confirming that legitimate code is legitimate.

\begin{table}[t]
\centering
\caption{Reviewer persona on the clean-aware code condition, primary model, 48 flawed and 48 clean items each. Only the assigned role changes.}
\label{tab:personas}
\footnotesize
\setlength{\tabcolsep}{1pt}
\begin{tabular*}{\columnwidth}{@{}l@{\extracolsep{\fill}}rrrr@{}}
\toprule
Persona & $R$ & FPR (95\% CI) & A3 & Conf. \\
\midrule
Auditor            & 100.0 & 0.0 (0.0--7.4)   & 87.5 & 0.953 \\
Quant reviewer     & 100.0 & 2.1 (0.4--10.9)  & 93.8 & 0.952 \\
Skeptical reviewer & 100.0 & 10.4 (4.5--22.2) & 93.8 & 0.928 \\
Risk manager       & 95.8  & 12.5 (5.9--24.7) & 91.7 & 0.911 \\
\bottomrule
\end{tabular*}
\end{table}

\paragraph{Persona moves calibration, not recall.}
Holding the model, the prompt mode, and the surface fixed, we vary only the reviewer role the system prompt assigns. Table~\ref{tab:personas} reports the four roles on the clean-aware code condition. Recall is flat at 100.0\% for three of them and 95.8\% for the fourth, while the clean-control rate moves from 0/48 to 6/48. The two roles that carry an adversarial connotation, \texttt{risk\_manager} and \texttt{skeptical\_reviewer}, account for 11 of the 12 false positives across the four roles, and \texttt{risk\_manager} is also the only role that misses a flawed item.

Mean self-reported confidence tracks the false-positive rate in the wrong direction: the roles that flag more clean controls report lower confidence (0.911 and 0.928 against 0.953 and 0.952), so a confidence filter would discard the flags of the better-calibrated roles first.

\paragraph{Surface matters, and not uniformly.}
Prose and code are not interchangeable inputs. Under the open prompt, prose recall exceeds code recall (81.2\% against 72.9\%) while prose false positives are less than half the code rate (37.5\% against 93.8\%). Under the closed prompt the recall gap closes to zero and the false-positive gap persists (2.1\% against 20.8\%). A natural-language description of a methodology therefore elicits a better-calibrated audit than the code that implements it, at least at these sample sizes, and the gap survives being told what to look for.

\section{Discussion}
\label{sec:discussion}

Under deterministic paired edits, prompt and model choice separate high flaw recall from clean-control calibration; this does not show that an LLM can validate a trading system in production. The takeaway for benchmark design is narrow and practical: backtest-auditor benchmarks should report flaw recall, clean false positives, evidence, fix, and confidence separately, because a single aggregate would have rated all four endpoints in Table~\ref{tab:models} as near-solved.

\section*{Limitations}

The sample is small, at 6 flawed examples per class, so per-class rates move in steps of 16.7 points and the Wilson intervals in Table~\ref{tab:models} are correspondingly wide. Many clean-aware results saturate at 100\% recall, which we treat as a finding rather than a solved task, since GPT-4o mini over-flags clean code controls despite perfect recall. The strict lexical scorer may under-credit unusual fixes; alias-only matches account for 56 of the correct responses, or 4.6\%, so the scorer's synonym table is a live sensitivity. Items are synthetic paired edits over a fixed set of strategy skeletons, not externally collected backtest notebooks, and every audit is a single temperature-0 call rather than a multi-turn review. A human baseline packet of 16 items was prepared but not completed, so no human reference rate is reported. A next validator should add semantic entailment for the fix column and externally sourced notebooks for the item pool.

\section*{Ethics Statement}

All backtests in the benchmark are synthetic, and no market data, broker record, or proprietary strategy is redistributed. The benchmark measures whether a model recognizes planted methodology flaws under matched controls. It does not certify any trading system, and a high score here is not evidence that a model can validate a real strategy or that any strategy is sound. We report clean-control false positives alongside recall because an auditing tool that over-flags imposes review cost and can crowd out the one submission that genuinely needs repair. The paper makes no claim about profitability and offers no investment advice.

\appendix

\section{Per-Class Recall Without the Taxonomy}
\label{app:classes}

Table~\ref{tab:classes} gives the primary auditor's open-prompt recall per flaw class on both surfaces, with the closed-prompt column for reference. The closed prompt recovers every class on both surfaces; the open prompt does not, and the shortfall concentrates in full-sample preprocessing and regime overfit.

\begin{table}[htbp]
\centering
\caption{Primary-auditor recall by flaw class, 6 flawed items per class. Open prompts omit the taxonomy; the closed column reaches 6/6 on every class and both surfaces.}
\label{tab:classes}
\small
\setlength{\tabcolsep}{3pt}
\begin{tabular}{@{}lccc@{}}
\toprule
Flaw class & \shortstack{open\\prose} & \shortstack{open\\code} & closed \\
\midrule
Look-ahead & 6/6 & 6/6 & 6/6 \\
Survivorship & 6/6 & 6/6 & 6/6 \\
Data snooping & 6/6 & 6/6 & 6/6 \\
Overlapping-label leakage & 6/6 & 5/6 & 6/6 \\
Unrealistic fills/costs & 5/6 & 5/6 & 6/6 \\
P-hacked stop/rule & 6/6 & 4/6 & 6/6 \\
Regime overfit & 4/6 & 2/6 & 6/6 \\
Full-sample preprocessing & 0/6 & 1/6 & 6/6 \\
\midrule
Total & 39/48 & 35/48 & 48/48 \\
\bottomrule
\end{tabular}
\end{table}

\newpage
\section{Recall Against False Positives by Surface}
\label{app:surface}

\begin{figure}[!ht]
\centering
\includegraphics[width=\columnwidth]{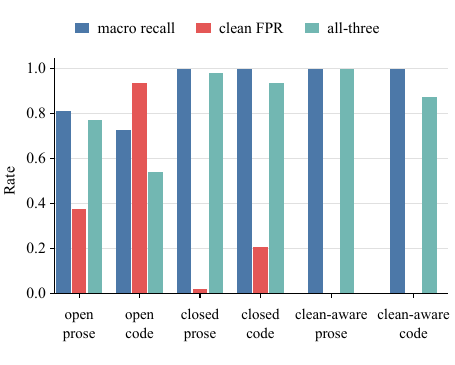}
\caption{Primary auditor recall against clean-control false positives, by surface and prompt mode.}
\label{fig:surfaces}
\end{figure}

Figure~\ref{fig:surfaces} plots the primary auditor's recall against its clean-control false-positive rate for each prompt mode and surface. The three prompt modes trace a path that moves up and to the left rather than along a single frontier: the clean-aware prompt gains recall over the open prompt and loses false positives at the same time, so the two are not being traded against each other.

\begin{thebibliography}{9}

\bibitem[White(2000)]{white2000}
H. White.
A reality check for data snooping.
\newblock \emph{Econometrica}, 68(5):1097--1126, 2000.

\bibitem[Harvey et~al.(2016)Harvey, Liu, and Zhu]{harvey2016}
C.~R. Harvey, Y. Liu, and H. Zhu.
\ldots\ and the cross-section of expected returns.
\newblock \emph{Review of Financial Studies}, 29(1):5--68, 2016.

\bibitem[Bailey et~al.(2017)]{bailey2014}
D.~H. Bailey et~al.
The probability of backtest overfitting.
\newblock \emph{Journal of Computational Finance}, 20(4):39--69, 2017.

\bibitem[Lopez~de~Prado(2018)]{lopez2018}
M. Lopez de Prado.
\emph{Advances in Financial Machine Learning}.
\newblock Wiley, 2018.

\bibitem[Wang et~al.(2026)]{wang2026}
Z. Wang et~al.
BacktestBench.
\newblock arXiv:2605.17937, 2026.

\bibitem[Li et~al.(2026)]{li2026}
W.~W. Li et~al.
Mitigating look-ahead bias in financial backtesting with LLMs.
\newblock arXiv:2605.24564, 2026.

\bibitem[Chaudhuri et~al.(2026)]{chaudhuri2026}
Y. Chaudhuri et~al.
E3 issue-level backtesting for automated research critique.
\newblock arXiv:2605.27072, 2026.

\bibitem[Ho et~al.(2026)]{ho2026}
S.-T. Ho et~al.
SoundnessBench.
\newblock arXiv:2605.30329, 2026.

\bibitem[Zurawicki et~al.(2026)]{zurawicki2026}
K. Zurawicki et~al.
PRAIB: LLM-assisted peer review benchmark.
\newblock arXiv:2605.29815, 2026.

\end{thebibliography}
\end{document}